\documentclass[
]{ceurart}

\usepackage{listings}
\begin{document}

\copyrightyear{2026}
\copyrightclause{Copyright for this paper by its authors.
  Use permitted under Creative Commons License Attribution 4.0
  International (CC BY 4.0).}

\conference{CMN'26: International Workshop on Computational Models of Narrative,
  June 8--10, 2026, Madrid, Spain}

% \title{Incremental Interpretation with Delayed Elaboration in Visual Narratives (Tentative Name,  8 and 16 pages, references included)}
\title{Distinguishing Revision and Delayed Elaboration in Incremental Narrative Interpretation}

% \tnotetext[1]{You can use this document as the template for preparing your
%   publication. We recommend using the latest version of the ceurart style.}

\author[1]{Yi-Chun Chen}[%
orcid=0009-0003-4035-9894,
email={rimi.chen@gs.ncku.edu.tw, ychen74@alumni.ncsu.edu},
url={https://www.csie.ncku.edu.tw/zh-hant/members/75},
]
\address[1]{National Cheng Kung University,
Tainan, Taiwan}

% \author[2]{Pablo Gervás}[%
% orcid=0000-0000-0000-0000,
% email=pgervas@ucm.es,
% url=https://www.ucm.es/itc/pablo-gervas,
% ]
% \address[2]{Universidad Complutense de Madrid, Madrid, Spain}

\begin{abstract}
Both human and AI systems that process narrative or long-form content operate incrementally: input is received over time, and internal representations must be updated accordingly. Incremental interpretation, therefore, depends not only on what is represented but also on how the representational state evolves under new evidence.

We distinguish two structurally different update operators that arise in narrative interpretation: revision-driven update and delayed elaboration. Revision-driven updates retract or replace previously committed structure in response to a contradiction and are therefore non-monotonic. Delayed elaboration, by contrast, refines initially underspecified elements through constraint addition without retracting prior commitments, yielding monotonic extension of the interpretive state. Although both operators may alter how earlier material is understood, they impose fundamentally different structural requirements on state transitions.

Using visual narratives as a diagnostic domain, we demonstrate how a structured narrative representation can explicitly separate committed from underspecified content and support both update operators during incremental construction. Through a worked example, we show how delayed elaboration enables monotonic refinement of interpretive state, while revision requires non-monotonic correction. We discuss the broader relevance of this structural distinction for incremental reasoning and hybrid symbolic–neural systems.

\end{abstract}

% 1. Introduction
% 2. Interpretation Update Dynamics: Revision vs. Elaboration
% 3. Visual Narrative as Diagnostic Domain
% 4. Minimal Incremental Representation Model
% 5. Worked Example
% 6. Discussion and Future Directions
% 7. Conclusion

% Introduction:
% Interpretation requires update operators.

% Related Work:
% Revision, Incrementality, Underspecification, Structured Models.

% Section 3:
% Define update operators.
% Show they exist in visual narrative.
% Show hierarchical representation can support them.
% Demonstrate with example.

% Discussion:
% If not distinguish them,
% overuse non-monotonic revision,
% destabilize explanation trace.


% Hybrid symbolic–neural systems \sep
% Interpretive commitment

\maketitle
% Table caption above the table
% Figure caption below the figure

\section{Introduction}
\label{sec:intro}

Many AI systems process input sequentially rather than receiving complete information in advance. Stories, conversations, and multimodal streams are interpreted incrementally: new observations are integrated into an evolving interpretive state \cite{garrod1999incrementality,cristea1997expectations}. In such settings, interpretation can be modeled as a state update process, as formalized in discourse theories \cite{kamp1988discourse,heim1982semantics} and implemented in incremental dialogue systems \cite{devault2011incremental}. Under incremental processing, representational design must specify not only what structures are encoded, but also how the interpretive state is updated over time.

A well-studied source of update is contradiction. When new input conflicts with previously committed structure, the system must retract or replace elements of the current state \cite{alchourron1985logic,doyle1978truth}. Such revision-driven updates are inherently non-monotonic: previously committed structure may no longer remain valid once new evidence is incorporated. Belief revision and truth maintenance frameworks provide formal accounts of this behavior.

However, not all updates arise from contradiction. In many sequential domains, early input introduces partial or underspecified structure that is later refined as additional evidence becomes available. Work on semantic underspecification and constraint-based representation demonstrates that incomplete structures can be maintained and progressively constrained without retraction \cite{egg2010semantic,pezzelle2023dealing,hutchinson2022underspecification}. In these cases, later updates add constraints to an existing structure rather than replacing it.

This paper distinguishes two representational update operators for incremental interpretation. \textbf{Revision-driven update} modifies previously committed structure in response to contradiction and therefore requires non-monotonic change \cite{alchourron1985logic,doyle1978truth}. \textbf{Delayed elaboration} refines previously underspecified components by adding constraints without removing earlier commitments, yielding monotonic extension of the interpretive state \cite{egg2010semantic,pezzelle2023dealing}. Although both operators may alter the interpretation of earlier input, they differ structurally in whether rollback is required.

We use \textbf{visual narratives} as a diagnostic domain to illustrate this distinction. Comics and manga present information panel by panel, making the incremental construction of structure explicit \cite{cohn2015getting,mccloud1993understanding,loschky2020scene}. Early panels often introduce entities or situations without fully specifying their relational roles, while later panels provide clarifying evidence. This sequential staging enables explicit examination of how an incremental representation can support monotonic elaboration without non-monotonic revision.

Our goal is not to propose a new interpretation engine. Instead, we analyze how an existing structured narrative representation can be constructed incrementally in a way that explicitly distinguishes between committed and underspecified components, thereby separating monotonic elaboration from non-monotonic revision.

\paragraph{Contributions.}
The contributions of this paper are threefold:
\begin{itemize}
  \item We define two structurally distinct update operators for incremental interpretation: revision-driven update (non-monotonic modification under contradiction) and delayed elaboration (monotonic refinement of underspecified structure).
  \item We demonstrate, through a worked visual narrative trace, how these operators correspond to different state transition patterns in a hierarchical graph representation.
  \item We discuss implications of explicitly distinguishing update operators for incremental reasoning systems and structured symbolic representations.
\end{itemize}

% \paragraph{Research Questions.}
% This paper addresses the following questions:
% \begin{itemize}
%     \item \textbf{RQ1:} What types of representational update are required during incremental narrative interpretation?
%     \item \textbf{RQ2:} How do revision-driven updates and delayed elaboration differ in terms of representational commitment and state transition?
%     \item \textbf{RQ3:} What are the consequences of conflating these update operators in structured narrative representations?
% \end{itemize}
% Goal: Motivate a general computational problem about incremental interpretation updates.

\section{Background and Related Work}
\label{sec:related}

We review prior work on belief revision, incremental interpretation, underspecification, and structured narrative representations. These areas show that interpretation changes over time and requires mechanisms for updating internal state. However, they do not clearly separate revision from refinement. Our work focuses on this distinction.

\subsection{Revision and Non-monotonic Update}

Revision-driven update has been widely studied in belief revision and non-monotonic reasoning. In these frameworks, new information may contradict previously accepted assumptions, requiring retraction or replacement of parts of the current knowledge state \cite{delgrande2018general,falakh2025agm}. Such updates are inherently non-monotonic: previously derived conclusions may no longer hold once new evidence is incorporated.

Belief revision theory formalizes this process through rationality constraints governing how a knowledge base should change under contradiction. These approaches emphasize consistency maintenance and controlled rollback of dependent inferences. Within narrative systems, revision has likewise been treated as an explicit operation when interpretive commitments must be corrected in light of later developments \cite{gervas2016integrating}.

This line of work establishes a revision-driven update as a well-defined mechanism for handling contradiction. However, it does not fully address cases in which later information refines earlier cues without invalidating them. Distinguishing such refinement from revision is central to our argument.

\subsection{Incremental Interpretation as State Update in Discourse and Narrative}

Interpretation in dialogue and narrative has long been modeled as an incremental process in which meaning is constructed online as input unfolds. In incremental dialogue systems, utterances are interpreted under streaming input conditions, and partial hypotheses must be maintained and updated as new material becomes available \cite{devault2011incremental}. These systems explicitly track evolving interpretive states rather than recomputing meaning from scratch at each step.

A similar perspective appears in computational models of narrative understanding. Narrative interpretation involves maintaining a representation that evolves over time, incorporating new events, updating temporal relations, and revising assumptions about event actuality or embedded contexts \cite{gervas2021model,gervas2024representing,gervas2025accounting}. Understanding is thus treated as a temporally extended process in which structural commitments accumulate and occasionally shift.

Research on visual narrative comprehension further reinforces this view. Structural accounts of comics describe how readers integrate panels incrementally, guided by narrative grammar and layout constraints \cite{cohn2013visual,cohn2014architecture}. Event-based models frame comprehension as the construction and updating of situation or event models over time \cite{loschky2020scene,brich2024construction}. Empirical evidence shows that readers generate inferences at specific transition points and adjust their internal representations as new cues are revealed \cite{cohn2015getting}. Computational approaches similarly implement stepwise integration of panels into an evolving representational state \cite{chen2021computational}.

Together, this body of work supports the view that narrative interpretation is a temporally extended state-update process. However, while prior research establishes that representations must evolve incrementally, it does not systematically distinguish between different types of update operations within that evolving state.

\subsection{Underspecification and Delayed Commitment}

In contrast to belief revision frameworks, which focus on retraction under contradiction, several traditions in computational semantics adopt underspecified representations that deliberately delay structural commitment. Rather than constructing a fully resolved interpretation at each step, these approaches maintain partially constrained representations that can be refined as additional information becomes available.

Underspecified semantic frameworks, such as Minimal Recursion Semantics \cite{copestake2005minimal} and underspecified discourse representation models \cite{schilder1998underspecified}, represent ambiguity and incomplete structure explicitly, allowing constraints to accumulate without forcing premature resolution. In these systems, refinement proceeds through monotonic constraint addition rather than retraction of prior commitments.

More recent work in multimodal and scene interpretation likewise emphasizes the need to represent partially specified meaning when perceptual or linguistic cues do not uniquely determine interpretation \cite{pezzelle2023dealing,hutchinson2022underspecification}. Here, underspecification is treated as a principled representational strategy rather than a temporary failure to decide.

This perspective provides a computational foundation for our notion of delayed elaboration. Early narrative cues may introduce placeholders that are intentionally incomplete; later context refines these structures without requiring non-monotonic revision.

\subsection{Structured Narrative Representations}

Computational models of narrative commonly employ structured representations to capture event relations, temporal order, causal dependencies, and character states. Multi-aspectual models argue that different structural layers are required to support interpretation and generation processes \cite{gervas2014need}. Event-centric and graph-based approaches encode narrative structure as networks of temporally and causally related events \cite{yan2023narrative,tang2022ngep,knez2023event}, making relational organization explicit and computationally accessible.

In visual narrative research, structured representations have been used to model panel transitions, narrative grammar, and event-level organization \cite{cohn2013visual,martens2020visual,chen2023panel}. These approaches demonstrate the importance of explicitly modeling structural relations across sequential inputs, particularly in settings where meaning emerges over time.

While such models make narrative organization explicit, they do not always distinguish between elements that are fully committed and those that remain intentionally underspecified. Consequently, refinement of partially specified content may be operationally indistinguishable from revision of prior commitments. Recent hierarchical knowledge graph approaches to visual narrative understanding \cite{chen2025hierarchical} model multi-level narrative structure but similarly do not explicitly differentiate between revision-driven and elaboration-driven updates.

The present work builds on structured narrative modeling traditions while focusing specifically on how representational design interacts with incremental update behavior. By distinguishing revision-driven updates from delayed elaboration within a structured representation, we make explicit the mechanisms that support monotonic refinement versus non-monotonic correction.

% Place related work here to justify the problem and situate the contribution.

%% 3 - 6 
% \section{Interpretation Update Dynamics: Revision vs.\ Delayed Elaboration} % Definition
% \section{Interpretation Change as a Process}
\section{Update Dynamics in Hierarchical Narrative Representation}
\label{sec:dynamics}

\subsection{Interpretive State and Update Operators}

We assume a hierarchical narrative representation in which panels are grouped into event segments, events, and macro-events through explicit structural relations. This layered structure encodes temporal order, participant roles, and event grouping within a unified graph \cite{chen2025hierarchical}. In the present study, we adopt this structural schema without modification. Rather than expanding its expressive scope, we examine how it behaves when constructed incrementally during interpretation.

Narrative understanding unfolds sequentially. As each panel is processed, the representation is extended to incorporate new observations and inferred structure. Let $S_t$ denote the interpretive state after processing the first $t$ panels of a narrative. $S_t$ consists of the partially constructed hierarchical graph at time $t$, including observed panel-level elements and any higher-level structure supported by available evidence. Because interpretation proceeds step by step, $S_t$ need not be complete. Some relations, role assignments, or event groupings may remain unresolved.

We distinguish between \emph{committed} and \emph{underspecified} components of $S_t$. A committed component is treated as resolved under the current interpretation. An underspecified component preserves an open role, provisional grouping, or incomplete classification that may later be refined as additional input becomes available. Within this incremental setting, we define two types of update.

%% may need block style for better presentation about definitions

\paragraph{Revision-driven update.}
A revision-driven update occurs when newly observed input contradicts previously committed structure. In such cases, elements of $S_t$ must be modified through deletion, relabeling, or structural replacement. If $S_{t+1}$ cannot be obtained from $S_t$ by monotonic extension, because previously committed nodes or edges must be altered, the update is non-monotonic. Revision-driven updates therefore involve correction of prior commitments.

\paragraph{Delayed elaboration.}
A delayed elaboration occurs when early cues introduce underspecified structure that is later refined without retracting prior commitments. In this case, $S_{t+1}$ extends $S_t$ by adding constraints, specifying roles, or completing structural relations while preserving all previously established nodes and edges. Such updates are monotonic: the graph grows through refinement rather than correction.

The distinction between these update types is representational rather than semantic. Both may change how earlier material is understood, but only revision requires rollback of committed structure. Delayed elaboration instead completes structure that was intentionally left open.

% \subsection{Structural properties of visual narratives}
% \begin{itemize}
%     \item Panel sequencing
%     \item Framing
%     \item Pacing
%     \item How visuals can introduce cues without specifying meaning
% \end{itemize}

% \subsection{Observed patterns of delayed elaboration (manga/comics)}
% \begin{itemize}
%     \item Summarize the patterns we observed
%     \begin{itemize}
%         \item early hint appears
%         \item meaning expanded later
%         \item no contradiction required
%         % Keep this part descriptive and modest
%     \end{itemize}
% \end{itemize}

% \subsection{Scope and selection notes}
% \begin{itemize}
%     \item State selected examples (a small set of Manga109 stories across genres)
%     \item Why (avoid adult content, reduce confounds)
%     \item This is exploratory evidence supporting the paper's modeling motivation.
% \end{itemize}

% \subsection{Visual Narratives as a Diagnostic Domain for Delayed Elaboration} %Why Visual Narratives Are Perfect Diagnostic Cases
\subsection{Visual Narratives as a Diagnostic Domain}
%Visual narratives unfold panel by panel. Information is literally withheld by layout.
Visual narratives provide a useful diagnostic domain for examining incremental update operators. In comics and manga, information is presented panel by panel, and each panel contributes partial structural evidence to an evolving interpretive state. Because panels are spatially segmented and temporally ordered, the sequential construction of structure is externally visible. This allows explicit tracing of how representational commitments are introduced and refined over time.

% Early panels are designed to be incomplete.
Importantly, visual narratives frequently rely on intentional underspecification. Early panels may introduce characters without specifying their relational roles, depict actions without establishing their causal relations, or present situations whose structural interpretation is resolved only later. Such staging is a common narrative technique for controlling information release. As a result, early interpretive states often contain underspecified components that are expected to be refined rather than revised.

% Close-up without context, Cropped scenes, Reaction before cause, Object shown before owner, Dialogue before speaker revealed
Visual framing further supports delayed elaboration. Cropped views, selective perspective, and staged revelation of contextual details postpone structural commitment. Rather than forcing immediate type assignment or relation labeling, the sequence allows incomplete components to persist until sufficient evidence is available. These properties make visual narratives particularly suitable for analyzing how underspecified structures are incrementally refined through monotonic extension.

\begin{figure}[t]
    \centering
    \setlength{\tabcolsep}{4pt} % horizontal spacing control
    \begin{tabular}{|ccc|c|}
    \hline
        \includegraphics[width=0.24\textwidth]{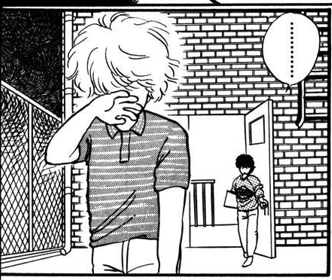} &
        \includegraphics[width=0.14\textwidth]{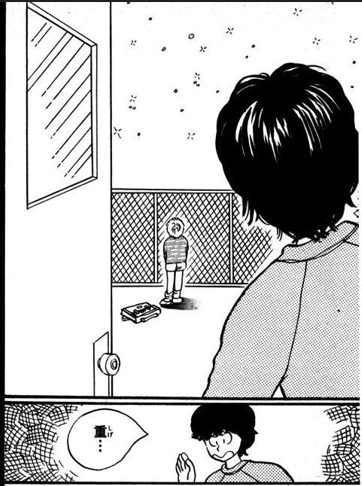} &
        \includegraphics[width=0.12\textwidth]{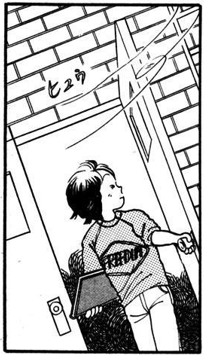} &
        \includegraphics[width=0.40\textwidth]{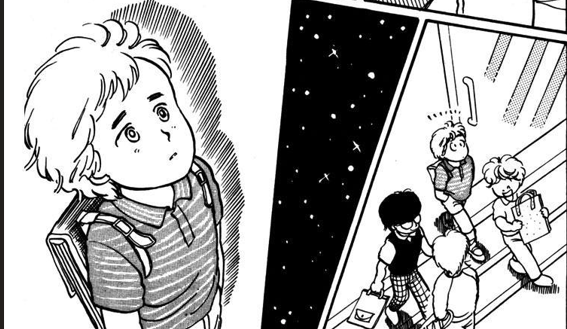} \\
        \hline
    \end{tabular}
    \caption{Selected panels from a visual narrative (reading direction: $\leftarrow$). 
    The panels are not contiguous; they are excerpted to illustrate how early cues introduce underspecified elements that are later refined without contradiction.}
    \label{fig:diagnostic}
\end{figure}

% Give an example
Figure~\ref{fig:diagnostic} illustrates a representative fragment. The panels (read right to left, following the original Japanese layout) first present entities and actions without fully specifying identity or relational structure. Later panels provide additional evidence that refines earlier components. No previously committed structure is removed; instead, earlier underspecified elements are completed through constraint addition.

% \begin{figure}[t]
%     \centering
%     \begin{minipage}{0.18\textwidth}
%         \includegraphics[width=\linewidth]{Images/Sample_004.png}
%     \end{minipage}
%     \hfill
%     \begin{minipage}{0.18\textwidth}
%         \includegraphics[width=\linewidth]{Images/Sample_003.png}
%     \end{minipage}
%     \hfill
%     \begin{minipage}{0.18\textwidth}
%         \includegraphics[width=\linewidth]{Images/Sample_002.png}
%     \end{minipage}
%     \hfill
%     \begin{minipage}{0.4\textwidth}
%         \includegraphics[width=\linewidth]{Images/Sample_001.png}
%     \end{minipage}
%     \caption{Selected panels from a visual narrative \textit{Reading direction:} $\leftarrow$. The panels are not contiguous; they are excerpted to illustrate how early cues introduce underspecified elements that are later refined without contradiction.}
%     \label{fig:diagnostic}
% \end{figure}

% clarify this is a case study to fix the scope
We use selected visual narrative examples not to propose a theory of comics comprehension, but to provide controlled cases for examining update operators in an incremental structured representation. These examples make visible the structural distinction between monotonic elaboration and non-monotonic revision in the evolution of the interpretive state.
% \subsection{Underlying representation}
% \begin{itemize}
%     \item Briefly summarize the existing visual narrative \/ hierarchical graph representation (nodes for events, entities, relations; multimodal attributes)
%     \item Cite previous prior work here.
%     \item State explicitly: this paper does not propose a new representation but extends previous ones.

% \end{itemize}

% \subsection{Incremental construction along the reading timeline}
% \begin{itemize}
%     \item Explain how the representation is built step-by-step as panels are consumed.
%     \item Emphasize that partial structures can exist without full semantic commitment.
% \end{itemize}

% \subsection{Representing underspecified cues}
% \begin{itemize}
%     \item Describe how visually or narratively salient cues are stored when their meaning is not yet determined.
%     \item This is the key extension: placeholders \/ partial nodes \/ open attributes.

% \end{itemize}

% \subsection{Elaboration without revision}
% \begin{itemize}
%     \item Show how later context expands or refines these structures.
%     \item Explicitly connect to truth revision idea of defeasible inference
%     \item Clarify that no contradiction-driven revision is involved here
% \end{itemize}
\subsection{Incremental Use of a Structural Representation for Delayed Elaboration}
\label{sec:delayed-elaboration}
% Incremental Construction of Hierarchical KG

% \begin{itemize}
%   \item The underlying visual narrative representation used in prior work is briefly summarized.
%   \item The incremental construction of this representation along the reading timeline is described.
%   \item Underspecified cues are represented as partial structures without immediate semantic commitment.
%   \item Later context elaborates these structures without invoking contradiction-driven revision.
% \end{itemize}

% How does the hierarchical representation actually behave incrementally when delayed elaboration occurs?

In this section, we describe how the hierarchical narrative representation can be constructed incrementally to support delayed elaboration as a monotonic update operator.

\paragraph{Structural components.}
The representation consists of panel-level nodes, event-segment nodes, event nodes, and macro-event nodes connected through structural relations such as temporal succession, containment, and participant-role links \cite{chen2025hierarchical}. Panel nodes encode observed visual elements, while higher-level nodes organize panels into structured event groupings. In the original formulation, this representation was constructed from fully annotated sequences. In the present setting, we instead consider incremental construction along the reading timeline, where the interpretive state $S_t$ is updated after each newly processed panel.

\paragraph{Panel-by-panel construction.}
As each panel $p_t$ is processed, new panel-level nodes and locally supported relations are added to the current interpretive state $S_t$. Only structure directly supported by the available evidence is committed at this stage. Higher-level groupings (event segments or event nodes) may be introduced when sufficient structural evidence accumulates, but they remain provisional if key roles or relations are not yet supported.
% need an example here

Crucially, not all structural components need to be fully specified when introduced. When participant identity, relational typing, or panel grouping remains uncertain, the representation encodes these components as \emph{underspecified} rather than committing to a fixed interpretation.

\begin{figure*}[t]
    \centering
    \setlength{\tabcolsep}{2pt} % control horizontal spacing
    % \begin{tabular}{|ccccc|ccc|}
    \begin{tabular}{|cccc|ccc|}
    \hline
        \includegraphics[width=0.16\textwidth]{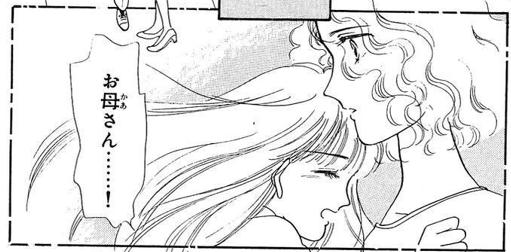} &
        \includegraphics[width=0.13\textwidth]{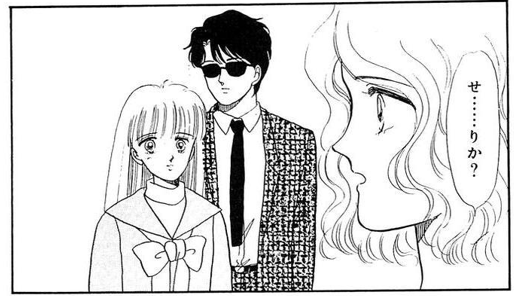} &
        \includegraphics[width=0.13\textwidth]{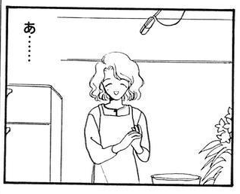} &
        \includegraphics[width=0.13\textwidth]{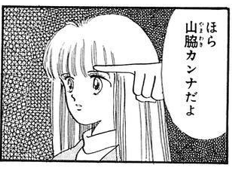} &
        \includegraphics[width=0.12\textwidth]{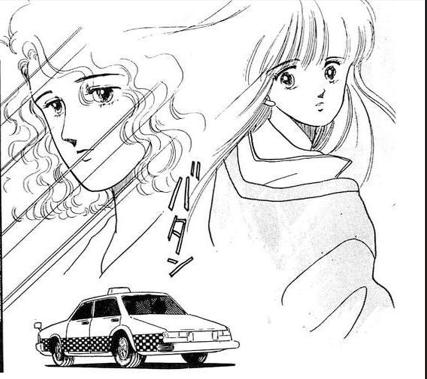} &
        \includegraphics[width=0.13\textwidth]{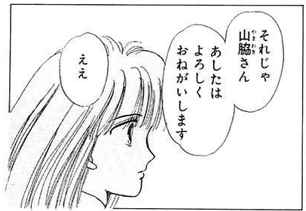} &
        \includegraphics[width=0.12\textwidth]{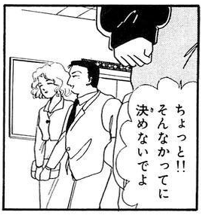} \\
    \hline
    \end{tabular}
    \caption{Excerpted seven-panel sequence used as a delayed-elaboration example (reading direction: $\leftarrow$). Early panels introduce a female character interacting with the protagonist without explicit relational typing. The final panel contains a dialogue-based identity reveal (``Mother!''), which licenses relational specification in the incremental trace.}
    \label{fig:graph_state_story_example}
\end{figure*}

\begin{figure}[t]
    \centering
    \includegraphics[width=0.55\linewidth]{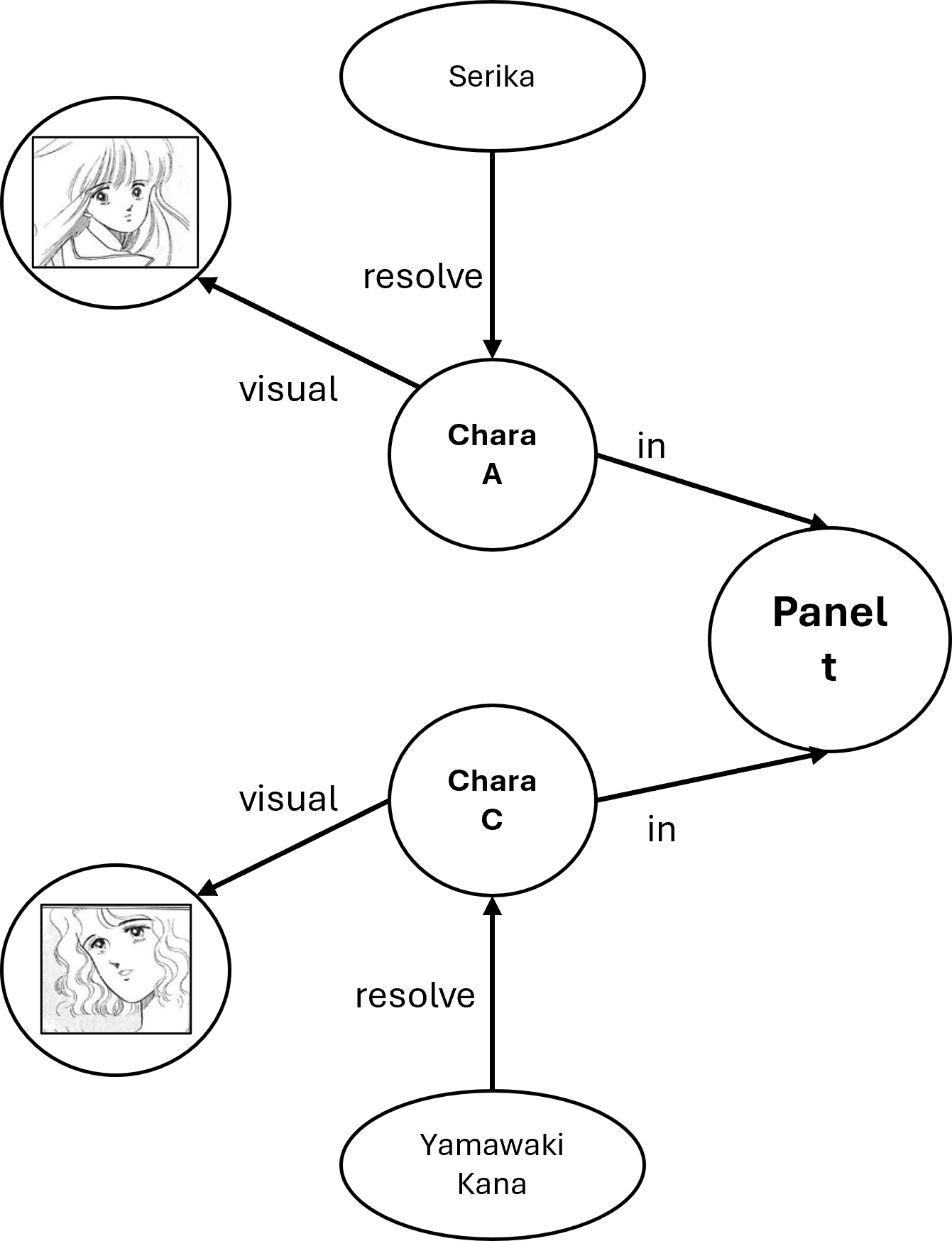}
    \caption{
    Early interpretive state $S_t$ for the narrative fragment shown in Figure~\ref{fig:graph_state_story_example}. 
    Entity nodes for the two characters are introduced and linked to the current panel. 
    Lexical labels (e.g., names) are resolved where supported by prior dialogue, 
    but no inter-entity role relation is yet committed. 
    The graph therefore contains structurally integrated entities with underspecified relational structure.
    }
    \label{fig:graph_state_early}
\end{figure}

We illustrate this with a delayed identity/role reveal. Figure~\ref{fig:graph_state_story_example} shows an excerpted panel sequence (reading direction: $\leftarrow$) in which a female character appears earlier without explicit relational typing. At this stage, the interpretive state contains a character node for the female but does not commit to a specific relational role with respect to the protagonist. The corresponding graph snapshot is shown in Figure~\ref{fig:graph_state_early}.
% Unresolved roles

\paragraph{Representing underspecification.}
Underspecification is implemented by leaving certain attributes or relational links untyped while maintaining their structural presence in the graph. For example, a character node may be introduced without a resolved participant-role relation in the current event, or an event segment may be created without committing to its final structural classification. These components remain available for later constraint addition. This design allows the interpretive state $S_t$ to contain structurally integrated but partially specified elements.
% need an example here

\paragraph{What revision would require in this representation.}
For completeness, we briefly note how revision-driven update would manifest within the same structural framework. If a previously committed node or relation were later contradicted, the interpretive state could not be extended monotonically. Instead, the graph would require modification through deletion, relabeling, or structural reattachment. For example, if an event grouping established at time $t$ were later invalidated by new evidence, the corresponding higher-level node would need to be replaced or reorganized. Such updates are non-monotonic and require explicit rollback of prior commitments.

In the present visual narrative example, however, the observed updates arise from refinement of underspecified components rather than contradiction of committed structure. Delayed elaboration therefore constitutes the dominant update operator in this diagnostic case.

\paragraph{Elaboration without revision.}
When later panels provide clarifying evidence, the representation is updated by adding constraints or completing previously underspecified roles. The resulting state $S_{t+k}$ is obtained by monotonic extension: no previously committed nodes or edges are removed. In the running example, later context confirms that the female character is the protagonist’s mother. Rather than modifying the earlier character node, the graph is refined by introducing the missing relational constraint. The earlier structure is preserved; its interpretation becomes more specific.

\begin{figure}[t]
    \centering
    \includegraphics[width=0.65\linewidth]{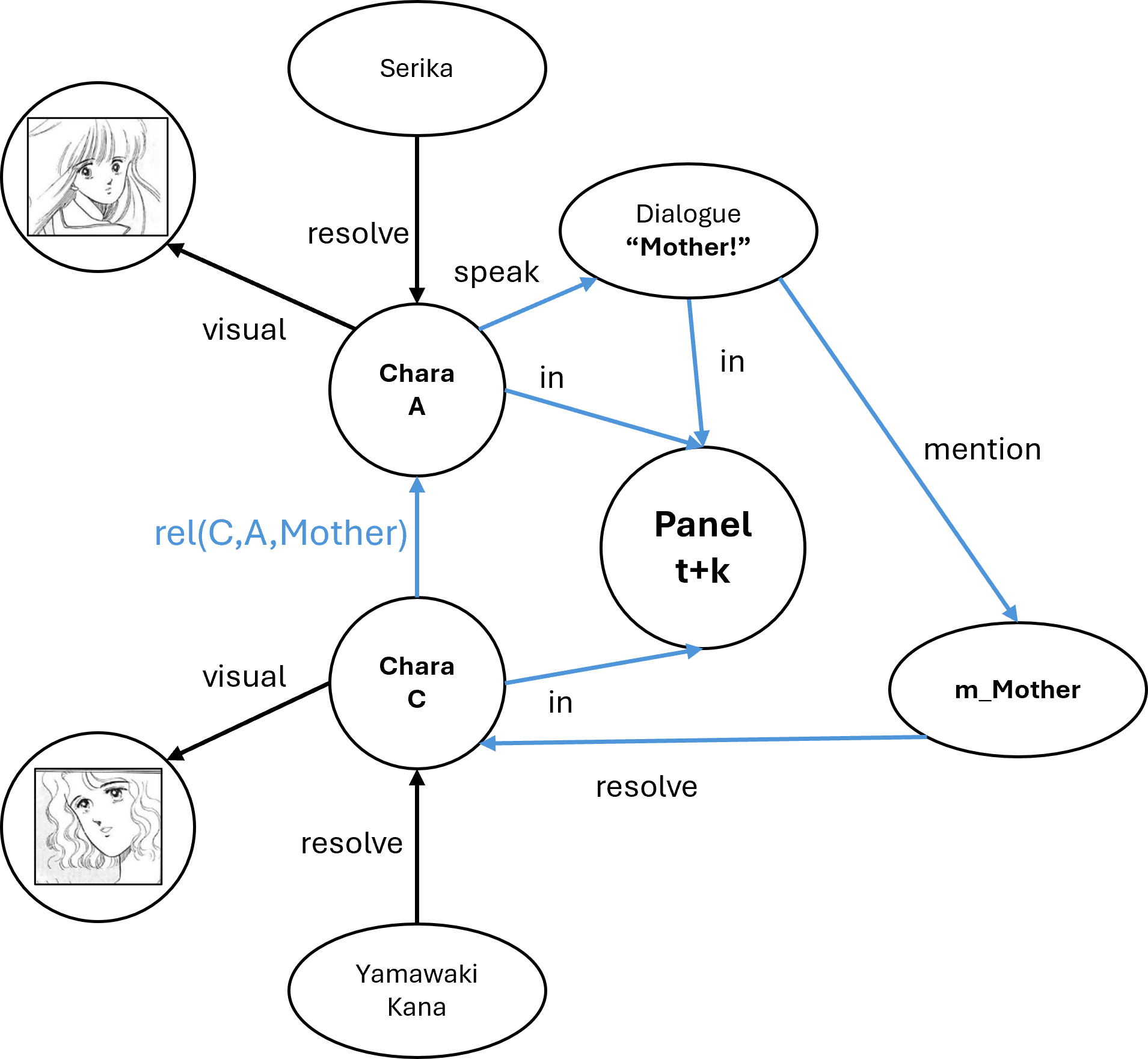}
    \caption{
    Later interpretive state $S_{t+k}$ for the same narrative fragment after the utterance ``Mother!'' is processed.
    A new mention node is introduced and resolved to the previously established entity node, 
    licensing the addition of the typed relation $\textsc{rel}(C, A, \text{Mother})$. 
    No previously committed nodes or edges are removed. 
    The update therefore constitutes monotonic elaboration rather than revision.
    }
    \label{fig:graph_state_late}
\end{figure}

The later refinement is shown in Figure~\ref{fig:graph_state_late}, where the relational constraint is added without altering prior structure.

% -----------------------------------------------------------------------------
% FIGURE PLACEHOLDER: "Later state" graph snapshot (elaborated)
% What to prepare:
%   - A graph rendering of S_{t+k} for the SAME story fragment as Figure~\ref{fig:graph_state_early}.
%   - Show that the previously underspecified slot is now resolved by ADDING:
%       (i) a new edge (e.g., role relation), and/or
%       (ii) a role label / attribute on the participant link,
%     while keeping all earlier committed nodes/edges unchanged.
%   - Optionally highlight the delta (new edges) with styling or callouts.
% -----------------------------------------------------------------------------

This incremental construction makes explicit the structural difference between monotonic elaboration and non-monotonic revision. Delayed elaboration extends the graph through constraint addition, whereas revision-driven update would require deletion or restructuring of previously committed components.

The absence of revision in this example reflects the staged information release characteristic of the selected sequence. The same representation would support revision if later input contradicted earlier commitments.

\subsection{Step-by-step incremental interpretation trace}
\label{sec:trace}

Section~\ref{sec:delayed-elaboration} illustrated delayed elaboration through early and late graph snapshots, $S_t$ and $S_{t+k}$. This section makes the incremental dynamics explicit by tracing the evolution of the interpretive state across seven panels of the same narrative sequence discussed above. The objective is to show how structural information is introduced, linked, and typed over time without requiring rollback of earlier commitments.

\paragraph{Representational assumptions.}
Each panel is represented as a multimodal graph containing character or entity nodes introduced visually, utterance nodes corresponding to dialogue, textual mention nodes extracted from dialogue expressions (e.g., proper names or address terms), and typed inter-entity relations added only when explicitly supported by visual or textual evidence. Characters persist across panels through identity links. Textual mention nodes may initially remain unresolved and can later be linked to specific entities once sufficient evidence becomes available. Typed inter-entity relations are introduced only when explicitly supported by dialogue or visual confirmation. The schema does not assume a fixed ontology of role types; relation labels are introduced incrementally when explicitly revealed in the evidence stream. 

For clarity of notation, we denote entity nodes by capital letters (e.g., $A, C$), utterance nodes by $u_i$, and textual mention nodes by $m_j$, where the subscript indexes the lexical content of the mention. Typed inter-entity relations are represented using predicate notation $\textsc{rel}(x,y,\ell)$, where $\ell$ is a relation label derived directly from dialogue.

Three update operations are distinguished. \textbf{ADD}: introduce nodes or edges directly supported by the current panel. \textbf{RESOLVE}: link a textual mention to a specific entity (identity or coreference resolution). \textbf{TYPE}: introduce a typed inter-entity relation once explicitly supported by evidence.

\paragraph{Trace overview.}
Table~\ref{tab:trace} summarizes the incremental updates. Each step corresponds to one panel. ``Graph delta'' reports only the structural changes introduced at that step.

\begin{table}[t]
\centering
\small
\setlength{\tabcolsep}{5pt}
\begin{tabular}{p{0.06\linewidth}| p{0.26\linewidth} |p{0.16\linewidth}| p{0.42\linewidth}}
\toprule
\textbf{Step} & \textbf{Panel cue} & \textbf{Operation} & \textbf{Graph delta (structural update)} \\
\midrule

$t_1$ & Two characters (C and D) walking; off-panel utterance by A: ``Wait—don’t decide that so casually.'' & ADD & Add character nodes $A$, $C$, $D$ (if not already present); add utterance node $u_1$ linked to speaker $A$; link visible characters to the panel node. \\
\hline
$t_2$ & Dialogue: ``Ms.~Yamawaki, we’re counting on you tomorrow.'' & ADD & Add utterance nodes $u_2, u_3$; create mention node $m_{\text{YamawakiMs}}$; link mention to utterance. No entity assignment is made at this stage. \\
\hline
$t_3$ & Character C shown leaving (no dialogue). & ADD & Add panel node; link $C$ to panel; optionally add action node (e.g., $\textsc{leave}(C)$) if action representation is supported in the panel-level graph. \\
\hline
$t_4$ & Character B points and says: ``That’s Yamawaki Kana.'' & ADD + RESOLVE & Add utterance node $u_4$; create mention node $m_{\text{YamawakiKana}}$; resolve $m_{\text{YamawakiKana}} \rightarrow C$; resolve earlier $m_{\text{YamawakiMs}} \rightarrow C$ (identity linking across panels). \\
\hline
$t_5$ & Character C shown alone (kitchen-like setting), no dialogue. & ADD & Add panel node; link $C$ to panel; add contextual nodes only if supported by the representational schema. \\
\hline
$t_6$ & Character C says hesitantly: ``Seri…ka.'' & ADD + RESOLVE & Add utterance node $u_5$; create mention node $m_{\text{Serika}}$; resolve $m_{\text{Serika}} \rightarrow A$. \\
\hline
$t_7$ & Character A hugs C and says: ``Mother!'' & ADD + RESOLVE + TYPE & Add utterance node $u_6$; create mention node $m_{\text{Mother}}$; resolve $m_{\text{Mother}} \rightarrow C$; introduce a typed inter-entity relation $\textsc{rel}(C,A,\text{Mother})$, where the relation label is derived directly from the dialogue expression. \\

\bottomrule
\end{tabular}
\caption{Incremental structural updates across seven panels of the sequence. Typed inter-entity relations are introduced only when explicitly supported by dialogue. No previously committed structure is removed in this trace.}
\label{tab:trace}
\end{table}

\paragraph{Delayed elaboration as late specification.}
Across $t_1 \ldots t_7$, the interpretive state evolves by additive extension. Earlier panels introduce characters and textual references without committing to any specific inter-entity role between them. The utterance ``Mother!'' at $t_7$ supports the introduction of a typed relation label between two already established entity nodes. The prior graph state was incomplete but not incorrect.

Formally, the update at each step can be characterized as:
\begin{equation}
S_{t+1} = S_t \cup \Delta_t,
\end{equation}
where $\Delta_t$ contains newly added nodes, identity resolutions, or newly typed relations. No deletion or rollback operation is required in this sequence.

\paragraph{Contrast with revision.}
If later evidence contradicted an established typed relation (e.g., explicitly denying the relation), the update would require removal or replacement of a previously committed edge:
\begin{equation}
S_{t+1} = (S_t \setminus R_t) \cup \Delta_t, \quad R_t \neq \emptyset.
\end{equation}
Such non-monotonic revision does not occur in the present trace. The identity reveal therefore constitutes delayed elaboration understood as late relational specification without rollback.

% % \section{Visual Narratives as a Diagnostic Domain for Delayed Elaboration}
% \section{Visual Narratives as a Diagnostic Domain}  % Dataset justification
% \label{sec:visual}
% % Goal: justify why visual narrative is a good environment to observe elaboration.
% \input{4_VNDomain}

% \section{Incremental Use of a Structural Representation for Delayed Elaboration} % Method
% \label{sec:model}
% \input{5_Model}

% \section{Example: Incremental Elaboration in a Visual Narrative} % Experiments (proof the hypothesis)
% \label{sec:example}
% % Goal: demonstrate the model behavior with a concrete example.
% \input{6_Example}

\section{Discussion: Implications Beyond Narrative and Toward Hybrid Systems}
\label{sec:discussion}

\label{sec:discussion}

\paragraph{Why distinguishing elaboration from revision matters.}
Incremental interpretation systems must decide when to introduce new structure and when to defer specification until additional evidence becomes available. Treating delayed elaboration as equivalent to revision conflates two different update behaviors: adding constraints to an incomplete structure versus modifying previously committed structure. By separating monotonic refinement from non-monotonic rollback, the update process becomes explicit and easier to analyze. This distinction clarifies when instability is due to genuine contradiction and when it is a consequence of early commitment. The contribution of this paper is therefore a structural characterization of update behavior in incremental interpretation, rather than a new ontology or task-specific performance improvement.

\paragraph{Structural consequences of distinguishing update types}

The incremental trace in Section~\ref{sec:trace} shows that late specification of a relation does not require removal of earlier structure when earlier states are represented as underspecified rather than prematurely typed. In this formulation, revision replaces previously asserted edges in response to contradiction, while delayed elaboration adds type information or constraints to existing nodes and links without deleting prior structure.

In the example sequence, earlier panels introduce entities and textual mentions without asserting a specific interpersonal relation. The final utterance “Mother!” introduces an explicit typed relation between two entities. Because no conflicting relation was asserted earlier, the update is monotonic. Distinguishing these cases makes the state transition behavior of the graph explicit and reduces unnecessary non-monotonic operations.

\paragraph{Interpretive stability and traceability}

Explicit underspecification also affects interpretive stability. When updates are monotonic, earlier graph states remain valid substructures of later states. This makes the evolution of the interpretation state traceable: each step can be categorized as node addition, edge addition, typing, or revision.

Such explicit update tracking is relevant for symbolic graph-based systems and architectures where intermediate reasoning states must be inspected or audited. Although the present work does not implement a separate explanation module, the structural separation of update types provides a clearer basis for analyzing how interpretation evolves over time.

\paragraph{Implications beyond narrative interpretation}

Visual narrative is used here because it frequently contains delayed specification of identity or role information. However, the structural distinction described is not specific to narrative content. Any incremental setting in which structure is constructed over streaming input may benefit from distinguishing refinement from correction.

Potential examples include dialogue interpretation, incremental knowledge graph construction, and multimodal event processing. We do not provide empirical evaluation in these domains, but the update framework presented here may serve as a conceptual basis for analyzing state transitions in other incremental reasoning systems.

\paragraph{Relation to hybrid symbolic--neural systems}

Neural sequence models typically encode commitment timing implicitly in their internal states. In contrast, symbolic graph representations make structural commitments explicit. Maintaining underspecified relations until explicit evidence is observed separates evidence accumulation from type assignment.

Although this work does not implement a neural component, the representation could serve as a structured interface in hybrid systems, where neural modules propose entity links or candidate relations and symbolic layers manage commitment, typing, and update classification.

\paragraph{Limitations}

This study is exploratory and conceptual. First, the analysis is based on a worked example rather than large-scale empirical evaluation. We do not measure frequency of revision versus elaboration across a corpus. Second, relation labels derived from dialogue remain open-ended and are not normalized to a predefined schema. The framework allows typed relations when explicitly indicated but does not enforce global constraints on relation types. Third, the trace demonstrates delayed elaboration but does not include a full empirical comparison with revision-heavy cases. Finally, automatic detection of underspecification versus contradiction is not implemented.

% Future work may include corpus-scale evaluation, relation normalization strategies, and integration with neural components for large-scale incremental interpretation.

\section{Future Directions}
\label{sec:future}
% Goal: show a path from workshop seed to AI conferencec trajectory.

% \subsection{From examples to systematic evaluation}
% \begin{itemize}
%     \item Propose evaluation dimensions
%     \item Trace stability
%     \item Explanation quality (unknown vs wrong) 
%     \item Reduced unnecessary revision
%     \item human alignment in incremental interpretation (optional user study).

% \end{itemize}

% \subsection{Benchmark construction for delayed elaboration}
% \begin{itemize}
%     \item Outline how to build a small benchmark
%     \item Annotate cues and elaboration points
%     \item Tasks: predict elaboration triggers \/ update operations \/ plausible elaborations
% \end{itemize}

% \subsection{Hybrid models for incremental interpretation}
% \begin{itemize}
%     \item A next-step system
%     \item LLM/VLM proposes hypotheses
%     \item Symbolic layer manages underspecification and elaboration
%     \item Compare against baselines once defined
% \end{itemize}

% \begin{itemize}
%   \item Possible evaluation criteria for elaboration-aware interpretation updates are outlined.
%   \item Directions for integrating neural prediction with symbolic elaboration mechanisms are discussed.
%   \item Extensions to other incremental interpretation domains are proposed.
% \end{itemize}
% %==================

% %% Future Work Potential Extension
% \cite{chen2025narrative}

% \section{Future Directions}

The present study establishes a structural distinction between revision-driven update and delayed elaboration within a controlled narrative setting. This distinction opens several research directions that extend beyond the illustrative example.

\paragraph{Formal update classification and algorithmic criteria.}
A first direction is to formalize structural criteria for classifying updates as revision or delayed elaboration. This would involve defining contradiction detection conditions, constraint-completion rules, and rollback triggers within graph-based representations. Such formalization could enable algorithmic update classification and complexity analysis of incremental state transitions.

\paragraph{Corpus-scale structural analysis.}
The current demonstration is example-driven. Extending the analysis to narrative corpora would allow measurement of revision frequency, elaboration frequency, and structural patterns of late specification. This would provide empirical grounding for the representational distinction and enable comparison against systems that do not explicitly separate update types.

\paragraph{Schema refinement and relation normalization.}
The framework currently introduces relation labels directly from dialogue without global normalization. Future work may investigate schema-level constraint modeling that preserves incremental flexibility while enabling cross-instance comparison and interoperability with existing knowledge graph resources. Such refinement would strengthen scalability and structural consistency.

\paragraph{Hybrid incremental architectures.}
The explicit separation between commitment, refinement, and rollback suggests a structured interface for hybrid systems. Neural components could propose candidate entity links or relation types under uncertainty, while a symbolic layer manages commitment timing and update classification. This separation may allow more controlled incremental reasoning in streaming multimodal settings.

\paragraph{Process-oriented evaluation metrics.}
Finally, the distinction motivates evaluation criteria that consider update dynamics in addition to final accuracy. Metrics such as rollback frequency, proportion of monotonic refinements, or state growth consistency may provide new dimensions for assessing incremental interpretation systems.

\section{Conclusion}
\label{sec:conclusion}
% \begin{itemize}
%     \item Restate
%     \begin{itemize}
%         \item revision vs delayed elaboration distinction
%         \item minimal representation supporting elaboration
%         \item worked example
%         \item broader relevance and next steps
%     \end{itemize}
% \end{itemize}

% \begin{itemize}
%   \item The distinction between revision and delayed elaboration is summarized.
%   \item The role of structural representations in supporting elaboration is reiterated.
%   \item The broader relevance of the proposed perspective is briefly restated.
% \end{itemize}

This paper examined how incremental interpretation can distinguish between two structurally different update behaviors: revision-driven update and delayed elaboration. While both arise during sequential processing of narrative input, they impose different requirements on representational state transitions.

Using visual narratives as a diagnostic domain, we showed that delayed elaboration can be modeled as monotonic refinement of explicitly underspecified structure, whereas revision requires non-monotonic modification of previously committed elements. By tracing the evolution of an incremental graph state across a panel sequence, we demonstrated how late specification of relational information can proceed without rollback when early commitments are represented conservatively.

The contribution of this work is not a new ontology, inference engine, or performance benchmark. Rather, it is a structural clarification of update behavior in incremental interpretation. Making the distinction between refinement and correction explicit, this study provides a clearer account of how representational commitments evolve over time and how rollback operations can be minimized when premature typing is avoided.

Although the examples are drawn from visual narrative, the distinction between update behaviour is not domain-specific. Any system that constructs structured representations under streaming input conditions may benefit from separating monotonic specification from non-monotonic revision at the representational level.

By framing incremental interpretation in terms of explicit update operators and state transitions, this work aims to contribute a representational perspective that can inform future developments in structured reasoning and hybrid symbolic–neural architectures.

%%
%% Define the bibliography file to be used
\bibliography{sample-ceur}

\appendix


% The sources for project are available via
% \begin{itemize}
% \item \href{https://github.com/RimiChen/2026_DelayedCommitment}{GitHub(currently private)}
% % \item \href{https://www.overleaf.com/project/5e76702c4acae70001d3bc87}{Overleaf},
% \end{itemize}

\section*{Acknowledgments}
The author used a large language model for language editing and formatting assistance. All conceptual development, formal definitions, examples, and analysis were authored and verified by the author.

\end{document}